\documentclass[sigconf]{aamas} 
\usepackage{algorithm,float}
\usepackage{lipsum}
\usepackage{xcolor}
\usepackage{soulutf8}
\usepackage{empheq}
\usepackage{algorithmicx}
\usepackage{algpseudocode}
\usepackage{amsmath}
\usepackage{tikz}
\usepackage{graphicx}
\usepackage{microtype}
\usepackage{nicefrac}
\usepackage{framed}
\usepackage{color,soul}

\usepackage{balance} 

\setcopyright{none}
\acmConference[ALA '26]%
{Proc.@ of the Adaptive and Learning Agents Workshop (ALA 2026)}%
{May 25 -- 26, 2026}%
{Paphos, Cyprus, https://alaworkshop2026.github.io/}%
{Aydeniz, Delgrange, Mohammedalamen, Yang (eds.)}%
\copyrightyear{2026}
\acmYear{2026}
\acmDOI{}
\acmPrice{}
\acmISBN{}
\acmSubmissionID{<<submission id>>}

\title[AAMAS-2026 Formatting Instructions]{CAPSULE: Control-Theoretic Action Perturbations for Safe Uncertainty-Aware Reinforcement Learning}

\author{Rahul Narava}
\affiliation{
  \institution{Indian Institute of Technology Ropar}
  \city{Ropar}
  \country{Punjab}}
\email{syam.21csz0018@iitrpr.ac.in}

\author{Siddharth Verma}
\affiliation{
  \institution{Indian Institute of Technology Ropar}
  \city{Ropar}
  \country{Punjab}}
\email{2022csb1126@iitrpr.ac.in}

\author{Ojas Jain}
\affiliation{
  \institution{Indian Institute of Technology Ropar}
  \city{Ropar}
  \country{Punjab}}
\email{2022csb1099@iitrpr.ac.in}

\author{Shashi Shekhar Jha}
\affiliation{
  \institution{Indian Institute of Technology Ropar}
  \city{Ropar}
  \country{Punjab}}
\email{shashi@iitrpr.ac.in}

\author{Mayank Shekhar Jha}
\affiliation{
  \institution{CRAN, Université de Lorraine}
  \city{Nancy}
  \country{France}}
\email{mayank-shekhar.jha@univ-lorraine.fr}

\begin{abstract}
Ensuring safe exploration in high-dimensional systems with unknown dynamics remains a significant challenge. Existing safe reinforcement learning methods often provide safety guarantees only in expectation, which can still lead to safety violations. Control-theoretic approaches, in contrast, offer hard constraint–based safety guarantees but typically assume access to known system dynamics or require accurate estimation of control-affine models. In this paper, we propose a safe reinforcement learning framework that learns a probabilistic control-affine dynamics model in an offline setting. The learned model is leveraged to explicitly construct control barrier functions (CBFs) that incorporate model uncertainty to provide conservative safety constraints. These CBF constraints are enforced through an online constraint-based action correction mechanism, enabling safe exploration without overly restricting task performance. Empirical evaluations on nonlinear, complex continuous-control benchmarks demonstrate that our approach achieves returns comparable to those of existing baselines while significantly reducing safety violations.
\end{abstract}


\newcommand{\BibTeX}{\rm B\kern-.05em{\sc i\kern-.025em b}\kern-.08em\TeX}

\begin{document}


\pagestyle{fancy}
\fancyhead{}


\maketitle 


\section{Introduction}
Reinforcement Learning (RL) studies the problem of sequential decision-making through interactions between an agent and its environment \cite{sutton2018reinforcement}. At each time step, the agent observes the current state, selects an action, and influences the subsequent evolution of the environment. In return, the environment provides feedback in the form of a scalar reward. Unlike supervised learning, RL does not rely on predefined labels; instead, the agent improves its behavior through trial-and-error interactions. A defining characteristic of RL is the presence of temporal dependencies, where actions affect not only immediate rewards but also future states and long-term outcomes. The objective of an RL agent is to learn a policy that maximizes the expected cumulative reward over time. RL methods have demonstrated remarkable success across a wide range of challenging domains \cite{trpo,ppo,sac}.

RL agents are driven by the objective of maximizing cumulative reward, which can lead to unsafe exploratory behaviors that incur high costs, damage the environment, or harm the agent itself. Without explicit safety considerations, this reward-driven optimization often encourages high-risk actions that push the agent into hazardous states and cause safety violations. This becomes especially critical in domains like autonomous driving, where unsafe exploration or policy deployment can endanger human lives. For example, a self-driving car must avoid collisions and respect speed limits rather than adopting aggressive maneuvers that cause accidents. Safe Reinforcement Learning (Safe RL) addresses these limitations by introducing mechanisms that enforce safety constraints during training, thereby keeping exploration within acceptable risk boundaries while still enabling effective policy learning. By embedding these constraints directly into the learning process, Safe RL provides a more reliable foundation for deploying RL systems in safety-critical real-world environments.

Exisiting Safe RL methods seeks to maximize reward while enforcing expected safety constraints. However, in highly safety-critical settings, such expectation-based constraints are often insufficient—an agent may still produce unsafe actions as long as they are rare on average. Because such frameworks cannot guarantee strict constraint satisfaction, this motivates looking toward control-theoretic methods within RL, which are specifically designed to provide stronger, more reliable safety guarantees.

In \cite{cheng,dob}, the authors enforce safety by solving a convex optimization that generates a corrective safety-preserving action on top of a base RL policy. Although effective, these approaches generally assume known system dynamics and are typically evaluated in low-dimensional or simplified settings. 
L1-MBRL \cite{L1} reports that enforcing a control-affine structure, where the learned dynamics depend linearly on the action. This can lead to unstable training and degraded performance. These issues become more prevalent in high-dimensional continuous-control environments, where the true dynamics are complex and highly nonlinear. As a consequence, ensuring safety guarantees under unknown dynamics remains a central challenge.


To address these limitations, this paper proposes \textbf{C}ontrol-theoretic \textbf{A}ction \textbf{P}erturbations for \textbf{S}afe \textbf{U}ncertainty-aware
Reinforcement \textbf{LE}arning dubbed as \textbf{CAPSULE}. This approach to explicitly learn a control-affine dynamics model suitable for online constraint-based safety filtering while avoiding the instability observed in prior online learning approaches. Our key idea is to decouple model learning from policy optimization through an offline pretraining phase that leverages a heteroscedastic probabilistic ensemble of dynamics models. This approach stabilizes the learning of the control-affine model, leveraging offline data before any online interaction begins. Offline pretraining mitigates the model bias 
reported in earlier works, yielding a much more stable model that is structurally compatible with our framework. The resulting model provides uncertainty estimates necessary for making safety-critical decisions in subsequent online learning.
Unlike \cite{cheng}, which relies on Bayesian models such as the Gaussian Process (GP) Model, i.e, effective but limited to low-data and low-dimensional regimes, our heteroscedastic ensemble provides scalable, PETS-style \cite{pet} uncertainty estimates suitable for modern continuous-control benchmarks. This uncertainty enables us to construct state-dependent safety margins and maintain a conservative approach. This allows our framework to enforce safety guarantees even in the presence of model error. Thus, the agent benefits from the exploration and performance while maintaining provable safety constraints derived from the learned dynamics and their uncertainty.
\\
\noindent Our contributions are of threefold:
\begin{itemize}
\item \textbf{Uncertainty Aware Control-Affine Probabilistic Ensemble Dynamics Model:} An offline-trained probabilistic dynamics model that maintains control-affine structure for higher dimensional environments along with heteroscedastic uncertainty estimate for safety guarantees.
\item \textbf{Online Safe RL via CBFs using learned models:} A hybrid method that pairs Reinforcement Learning with a CBF based safety correction using a pre-trained model allowing for hard-constraint satisfaction which is required in safety critical settings.
\item \textbf{Evaluation on Complex Continuous-Control Domains:} Experimentation over complex settings such as Safety Gymnasium (MuJoCo based) showing improved performance and fewer safety violations compared to existing safe RL approaches.
\end{itemize}
\section{Background}
\subsection{Markov Decision Process (MDP)} MDP is formulated by the tuple ${M}\!=<\!S,A,{R},\mathcal{P},\gamma\!>$ where $S$ is the state space, $A$ is the action space and $R\!:\!(S\times{A})\!\rightarrow\![r_{\text{min}}, r_{\text{max}}]$ is the reward function such that $ r_{\text{min}}, r_{\text{max}} \in \mathbb{R}$. Further, $\mathcal{P}\!:\!(S\times{A}\times{S})\rightarrow[0,1]$ captures the transition dynamics of the environment i.e $\mathcal{P} (\boldsymbol{\cdot} |s,a)$ is a probability distribution over the state space that outputs the probability of agent traversing into next state $s'$ given current state $s$ and action taken $a$. Discounting parameter is given by $\gamma \in [0,1)$.

A policy $\pi:(S\times A)\rightarrow[0,1]$ is a function that outputs the probability of taking an action $a$ in the state $s$ i.e a mapping $\pi(\cdot\!\mid\!s)$ from states to probability distribution over actions. The objective of a reinforcement learning (RL) agent is to learn a policy $\pi^*$ that maximizes the expected discounted return given by
$\mathbb{E}_\pi \left[ \sum_{t=0}^{\infty} \gamma^{t} R(s_t, a_t)\right]$. The state-value function, or simply the value function, $V^\pi(s)$, under a policy $\pi$, is defined as the expected return when starting from state $s$ and following policy $\pi$ thereafter:
\begin{equation}
    V^\pi(s) = \mathbb{E}_\pi \left[ \sum_{t=0}^{\infty} \gamma^{t} R_{t} \mid S_0 = s \right]
\end{equation}
The Q-value function, or action-value function, $ Q^\pi(s, a)$ under a policy $ \pi$, is defined as the expected return (cumulative discounted reward) starting from state \( s \), taking action \( a \), and thereafter following policy \( \pi \):
\begin{equation}
    Q^\pi(s, a) = \mathbb{E}_\pi \left[ \sum_{t=0}^{\infty} \gamma^{t} R_{t} \mid S_0 = s, A_0 = a \right]
\end{equation}

\paragraph{Trust Region Policy Optimization} TRPO is an on-policy actor-critic algorithm developed to improve the stability and reliability of policy optimization.
Standard policy gradients can take large, unconstrained steps in parameter space, causing abrupt changes in the policy. TRPO mitigates this issue by restricting each policy update within a trust region, defined by a Kullback–Leibler (KL) divergence constraint, ensuring that the new policy does not deviate excessively from the previous one and thereby achieving more stable and monotonic improvement.
Formally, TRPO maximizes the expected advantage under the old policy, subject to a constraint on the Kullback–Leibler (KL) divergence between successive policies:
\begin{equation}
\begin{aligned}
\max_{\theta} \quad & 
\mathbb{E}_{s,a \sim \pi_{\text{old}}} 
\left[ 
\frac{\pi_{\theta}(a|s)}{\pi_{\text{old}}(a|s)} 
A^{\pi_{\text{old}}}(s,a)
\right] \\
\text{subject to} \quad & 
\mathbb{E}_{s \sim \pi_{\text{old}}}
\left[
D_{\text{KL}}\!\left(
\pi_{\text{old}}(\cdot|s) \,\|\, \pi_{\theta}(\cdot|s)
\right)
\right]
\le \delta
\end{aligned}
\end{equation}
Here, $ A^{\pi_{\text{old}}}(s,a)$ denotes the advantage function, and $ \delta $ controls the maximum policy divergence.  
By enforcing this KL constraint, TRPO achieves monotonic improvement guarantees and more stable policy updates compared to vanilla policy gradient or standard actor-critic methods.

\paragraph{Constrained Markov Decision Process (CMDP)} is an extension of an MDP framework with constraint set $C$. Where constraint set C is $C=\{c_{i},b_{i}\}_{i=1}^{n}$ 
In the context of safe reinforcement learning, the goal is to optimize the discounted return ${\max}~\mathbb{E}[G_{t}]$  where $G_{t}=\sum_{k=0}^{\infty}\gamma^{k}r_{t+k+1}$ while also ensuring safety by avoiding violations or risky actions.


The objective in CMDP is to find a policy $\pi$ that maximizes the expected return while abiding by the constraints on expected cumulative cost under the threshold $b$.
\begin{equation}
\begin{array}{ll}
\max_{\pi} \quad & J_r(\pi) = \mathbb{E}_{\tau \sim \pi} \left[ \sum_{t=0}^{\infty} \gamma^t r(s_t, a_t) \right] \\
\text{s.t.} \quad & J_{c_{i}}(\pi) = \mathbb{E}_{\tau \sim \pi} \left[ \sum_{t=0}^{\infty} \gamma^t c(s_t, a_t) \right] \leq b_{i}
\end{array}
\label{cmdpobj}
\end{equation}

\paragraph{Control Barrier Functions} Consider a nonlinear control-affine system with state $s \in \mathbb{R}^n$ and action $a \in \mathbb{R}^{m}$, whose dynamics are given by $\dot{s} = f(s)+g(s)a$,
where $f : \mathbb{R}^n \to \mathbb{R}^n$ and $g : \mathbb{R}^n \to \mathbb{R}^{n \times m}$ are locally Lipschitz functions.

Safety constraints are specified by a continuously differentiable function $h : \mathbb{R}^n \to \mathbb{R}$,
which induces the safe set $\mathcal{C}: \{s \in \mathbb{R}^n | h(s)>0\} $.
The function h is called a Control Barrier function(CBF) , for all s in a neighbourhood of $\mathcal{C}$, the following condition holds:
\begin{equation}
    h(s_{t+1}) \ge (1 - \alpha)\, h(s_t), \quad \alpha \in (0,1]
\end{equation}
\begin{equation}
    h(f(s_t)+g(s_t)a)\ge (1 - \alpha)\, h(s_t)
\end{equation}

This inequality defines a state-dependent constraint on the action $a$, ensuring that the safe set $\mathcal{C}$ is forward invariant. If the system starts within $\mathcal{C}$, any action satisfying the above condition guarantees that the state remains safe for all future time. 

In practice, the action that is produced by a policy $a=\pi(s)$ can be modified by projecting it onto the set of admissible actions defined by the CBF constraint, typically via a Quadratic Program (QP). This allows CBFs to be integrated with learning-based or optimal controllers while preserving safety by construction.
\begin{figure}
    \centering
    \includegraphics[width=\linewidth]{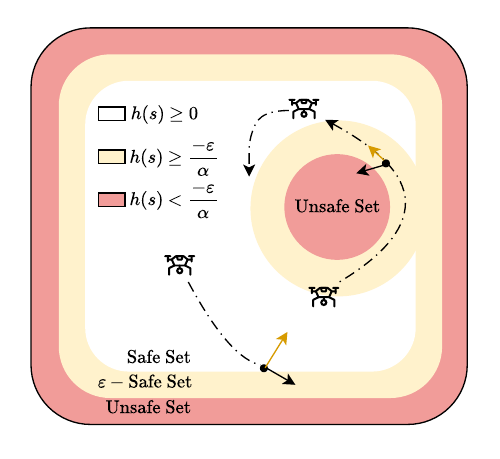}
    \caption{Illustration of the safe, $\epsilon$-safe, and unsafe sets induced by a control barrier function.}
    \label{fig:setdef}
\end{figure}
\paragraph{Forward Invariance} refers to the property of a set where, if the agent starts from that set, it remains in the same set for all future time under the given dynamics and control input. 
As we define $\mathcal{C}$ is a safe set, the set $\mathcal{C}$ is forward invariant under a control law $a$ if:
$s(0)\in \mathcal{C} \implies x(t)\in \mathcal{C} ~~\forall t\geq 0$.
i.e, the trajectory never leaves the safe set once it starts inside it.

CBFs are explicitly designed to enforce safety constraints over time. To achieve the safety, CBFs enforces forward invariance of the set.

\paragraph{Set Definitions in CBFs}
If $h : \mathbb{R}^n \to \mathbb{R}$ be a continuously differentiable function used to define the safe set. As illustrated in Figure \ref{fig:setdef}, the following are the set definitions of CBFs:
\begin{itemize}
    \item \textbf{Safe Set}, $\mathcal{C} \triangleq \{ x \in \mathbb{R}^n \mid h(x) \ge 0 \}$.
    \item \textbf{Interior} of the safe set, $\mathrm{Int}(\mathcal{C}) = \{ x \mid h(x) > 0 \}$.
    \item \textbf{Boundary} of the safe set, $\partial \mathcal{C} = \{ x \mid h(x) = 0 \}$.
    \item \textbf{$\epsilon$-Safe Set}, $\mathcal{C}^\epsilon \triangleq \{ x \in \mathbb{R}^n \mid h(x) \ge \frac{-\epsilon}{\alpha} \}$.
    \item \textbf{Unsafe set}, $\mathcal{C}^u \triangleq \{ x \in \mathbb{R}^n \mid h(x) < \frac{-\epsilon}{\alpha} \}$
\end{itemize}
any control input satisfying the CBF
constraint renders the safe set
$\mathcal{C} \triangleq \{x \mid h(x)\ge 0\}$ forward invariant.
Moreover, enforcing the relaxed CBF condition corresponding to the
$\epsilon$-safe set
$\mathcal{C}^{\epsilon} = \{x \mid h(x)\ge -\epsilon/\alpha\}$
guarantees forward invariance of $\mathcal{C}^{\epsilon}$, ensuring that trajectories
initialized within $\mathcal{C}^{\epsilon}$ do not enter the unsafe set
$\mathcal{C}^{u}$.

\section{Proposed Method}

\subsection{Offline Pretraining}
Learning the dynamics model online during policy optimization leads to the controller using unstable dynamics and can pose issues like policy divergence \cite{L1} and unsafe exploration. To enable safe policy learning without access to explicit system dynamics, we train a dynamics model in an offline manner prior to actual online policy learning. 
We define the underlying system as a parameterized control-affine probabilistic model:
    \begin{equation}
    \Delta s_t \mid s_t, a_t
    \sim \mathcal{N}\!\left( \mu_\theta(s_t,a_t),\;
    \sigma_\theta^2(s_t) \right)
    \label{eq:dist}
    \end{equation}
with control-affine mean:
\begin{equation}
\mu_\theta(s_t,a_t) = \hat f_\theta(s_t) + \hat g_\theta(s_t)a_t
\end{equation}
and state-dependent aleatoric uncertainty \(\sigma_\theta(s)\), where \(\hat f : \mathbb{R}^{|S|} \rightarrow \mathbb{R}^{|S|}\) models the drift dynamics, and \(\hat g : \mathbb{R}^{|S|} \rightarrow \mathbb{R}^{{|S|} \times {|A|}}\) characterizes the effect of action. We avoid $\sigma_\theta(s,a)$ since it complicates learning and yields poor uncertainty estimates from limited samples.
The choice of predicting the change in state rather than the next state is to stabilize training since the model learns to predict only the local drift dynamics, which are typically translation-invariant and do not depend on the absolute placement of the state in space.
For learning, we utilize the loss function 
\begin{equation}
\mathcal{L}(\theta)
=
-\sum_{n=1}^{N}
\log
\mathcal{N}\!\big(
\Delta s_n \,\big|\,
\mu_\theta(s_n,a_n),
\sigma_\theta(s_n)
\big)
\end{equation}
which then on simplifying becomes
\begin{equation}
\mathcal{L}(\theta)
=
\frac{1}{N}
\sum_{n=1}^{N}
\left(
\frac{\|\Delta s_n - \mu_\theta(s_n,a_n)\|^{2}}
{2\sigma_\theta(s_n)^{2}}
+
\log\!\big(\sigma_\theta(s_n)^{2}\big)
\right)
\label{eq:dyn_loss}
\end{equation}

We incorporate model ensembling to account for epistemic uncertainty in the dynamics. Note that this doesn't affect the control-affine structure that we previously defined. The final formulation thus comes out as given:
\begin{equation}
\bar\mu(s_t,a_t)
= \bar f(s_t)
+ \bar g(s_t)\, a_t,\bar\sigma(s_t)
\end{equation}
where
\[
\bar f_{\text{}}(s_t)
= \frac{1}{K}\sum_{k=1}^{K} \hat f_{\theta_{k}}(s_t),
\qquad
\bar g_{\text{}}(s_t)
= \frac{1}{K}\sum_{k=1}^{K} \hat g_{\theta_{k}}(s_t),
\]
and the ensemble-averaged aleatoric uncertainty is
\[
\bar \sigma(s_t)
= \frac{1}{K}\sum_{k=1}^{K} \sigma_{\theta_{k}}(s_t).
\]
This formulation helps us to capture total uncertainty in the learned model \cite{pet} which can then be utilized for uncertainty-aware safe exploration. To learn these models we use an offline dataset \(D_{\text{offline}}\) composed of transitions of form \((s_t, a_t, s_{t+1})\) collected from rollouts or demonstrations. The dataset is supposed to provide sufficient state coverage in order for the model to generalize better. The pretrained dynamics model is later used to construct the safety barrier used alongside online reinforcement learning.

\begin{algorithm}[h]
\caption{Offline Pretraining}
\label{alg:offline_pretraining}
\begin{algorithmic}[1]

\State \textbf{Input:} Offline dataset $\mathcal{D}_{\text{offline}}$, ensemble size $K$, pretraining steps $P$
\State Initialize ensemble parameters 
$\{\theta_{1}, \theta_{2}, \ldots, \theta_{K}\}$ for  
$\{\hat f_{\theta_k}, \hat g_{\theta_k}, \sigma_{\theta_k}\}$

\For{$i = 1, 2, \ldots, P$}

    \For{$k = 1, 2, \ldots, K$}
        \State Sample mini-batch $\{(s_n, a_n, s_{n+1})\} \in D_{\text{offline}}$
        \State Compute $\Delta s_n = s_{n+1} - s_n$
        \State Update $\theta_k$ using $\nabla_{\theta_k}\mathcal{L}(\theta_k)$ in \eqref{eq:dyn_loss}
    \EndFor

\EndFor

\State Compute ensemble averages $\bar f(\cdot), \bar g(\cdot), \bar \sigma(\cdot)$

\State \textbf{Output:} Trained ensemble dynamics model 
$\{\theta_1,\ldots,\theta_K\}$ and ensemble mean functions 
$(\bar f, \bar g, \bar\sigma)$.

\end{algorithmic}
\end{algorithm}

\subsection{Safety Barrier Construction}










We employ CBFs to enforce safety during online learning.
These encode the safety constraints that must be satisfied while the RL agent is exploring the environment. The resolution of CBFs is relatively less computationally expensive and is solved using QP solvers as demonstrated in \cite{ames, cheng, dob}. These are compatible with the control-affine dynamics model and can be used to solve for safety-corrective action update.\\
Computation of the CBF requires knowledge of the dynamics model to compute the next state and, subsequently, the corrective action. To encounter this, we employ our previously trained ensemble model and use it to predict the next state. Directly using the mean prediction can lead to over-estimation of safety, as it does not account for the uncertainty in model prediction. To mitigate this issue, we also consider the heteroscedastic uncertainty estimate provided by our probabilistic model. Leveraging the Gaussian modelling assumption of the $\Delta s_t$, we take a lower bound on the model error and ensure a conservative safety guarantee. 
\[
s_{t+1} \mid s_t, a_t
\sim \mathcal{N}\!\left( \mu_\theta(s_t,a_t) + s_t,\;
\sigma_\theta^2(s_t) \right)
\]
\begin{equation}
\implies s_{t+1} \in \mu_\theta(s_t,a_t) + s_t \pm p_\delta \sigma_\theta(s_t)
\label{eq:nextstatedist}
\end{equation}

Where $p_\delta$ defines the confidence interval on $s_{t+1}$ and ensures $s_{t+1}$ remains in the required range with a probability of $(1-\delta)$. This is used in the CBF formulation in the following way
{\small\begin{equation}
\sup_{a_t \in A} \Big[
    h\!\left(s_t + \bar f(s_t) + \bar g(s_t)a_t
        - p_\delta |\bar\sigma(s_t)| \right)
    - h(s_t)
\Big]
\ge
-\alpha\, h(s_t)
\end{equation}
}
Also, exact feasibility may not always be possible, for example, if there is no safe next state possible for the agent, we allow a non-negative slack $\epsilon$ to permit controlled constraint relaxation to still be in an $\epsilon$-safe set. This provides a graceful degradation of the safety constraints. At each step, the safe control action is obtained by solving 
\[
\begin{aligned}
a^*
&= \arg\min_{a,\epsilon}\ a^Ta + k\epsilon^2 \\[4pt]
\text{s.t. }\;
& h\!\left(s_t + \bar f(s_t) + \bar g(s_t)a
 - p_\delta|\bar\sigma(s_t)|\right)
 - h(s_t)\\
&\qquad
\ge -\alpha h(s_t) - \epsilon,\;\;\; \epsilon \ge 0
\end{aligned}
\]

where \(k >> 0\) penalizes slack usage and regulates how strictly the barrier constraint is enforced. When there is a valid action that satisfies the constraint, the optimal solution has \(\epsilon = 0\). This formulation allows the learned CBF constraint to be enforced consistently during online RL Policy training while accommodating model approximation error and control limitations. We leverage the pre-trained dynamics model along with the safety barrier constraints in online policy optimization.

\subsection{Online Policy Optimization}
\begin{algorithm}[t]
\caption{CAPSULE} \label{alg:Alg2}
\begin{algorithmic}[1]
\State \textbf{Input}: pretrained ensemble models $\bar{f}, \bar{g}, \bar{\sigma}$
\State \textbf{Initialize}: policy buffer $\text{B}$, policy network $\pi_{\theta}^{RL}$ , Compensator $\pi^{bar{}}_{\phi_k}$, Compensator buffer $\hat{\text{B}}$
\For{$k = 1,2,3....\text{Epochs}$}
    \State Initialize state $s_0 \sim \rho$
    \For{$t = 1, \dots, T$}
        \State Sample $a_t^{\text{RL}} \sim \pi^{\text{RL}}_k(s_t)$
        \State Sample $a_t^{\text{bar}} \sim \pi^{\text{bar}}_k(s_t)$
        \State Solve for $a_t^{\text{CBF}}(s_t,a_t^{\text{RL}}+a_t^\text{bar}$) using ~\eqref{qp}
        \State Execute $a_t$ from \eqref{safeaction} to get $(s_{t+1}, r_t)$
        \State Store $(s_t, a_t^{\mathrm{CBF}} + a_t^{bar})$ in $   \hat{\text{B}}$
        \State Store $(s_t, a_t, s_{t+1}, r_t)$ in $\text{B}$
    \EndFor
    \State Train RL policy $\pi_k^{\text{RL}}$ using $\text{B}$ to get $\theta_{k+1}$
    \State Train Compensator $\pi^{bar{}}_k$ using $\hat{\text{B}}$ to get $\phi_{k+1}$
\EndFor
\State \Return $\pi_{\theta}^{\text{RL}}, \pi^{bar{}}_{\phi} $
\end{algorithmic}
\end{algorithm}

 In this section, we outline the online training phase, where the agent interacts with the environment. During this stage, the objective is to optimize the task reward while ensuring that all the executed actions satisfy the safety constraints. Rather than executing the raw actions produced by the RL policy, the system employs a corrective control structure that modifies actions when necessary to maintain safety.

After each training epoch, the policy network   $\pi_{\theta}^{\text{RL}}$ is updated using a standard RL algorithm such as SAC, TRPO, or PPO. In this work, we employ TRPO as the base RL algorithm, yielding an updated task-driven controller $\pi_{\theta_{k+1}}^{\text{RL}}$.

However, executing $\pi_{\theta_k}^{\text{RL}}$ directly will lead to constraint violations. To mitigate this, our method incorporates two additional corrective components: (1) a safety compensation term $a_t^{bar}$ inspired from \cite{cheng}, that incrementally approximates the accumulated history of past CBF corrections, and (2) a real-time CBF controller $a_t^{\text{CBF}}$, computed by solving a constrained quadratic program using the pre-trained dynamics model.
At each timestep \(t\), the system first samples a candidate action \(a_t^{\mathrm{RL}}\) from the current policy. The compensator update \(a_t^{\mathrm{bar}}\) that accounts for safety history is added as correction to \(a_t^{\mathrm{RL}}\). Eventually, the Compensator estimates the safety provided by the CBF corrections and reduces reliance on CBF Controller. Finally, the CBF controller minimally adjusts this action to enforce forward invariance of the safe set, or if not possible, then in the $\epsilon$-safe set. The resulting executed action is the composition:
\begin{equation}
        a_t^{\text{safe}} = a_t^{\mathrm{RL}} + a_t^{bar} + a_t^{\mathrm{CBF}}(s_t,\;a_t^{\mathrm{RL}} + a_t^{bar})
        \label{safeaction}
\end{equation}

The resulting optimization equation to solve the CBF then becomes
\begin{equation}
\begin{aligned}
a_t^{\mathrm{{CBF}}}
&= \arg\min_{a,\epsilon}\ a^Ta + k\epsilon^2 \\
\text{s.t. }\;
& h\left(s_t + \bar f(s_t) + \bar g(s_t)(a_t^{\mathrm{RL}} + a_t^{bar}+a)
 - p_\delta|\bar\sigma(s_t)|\right)
 \\
&\qquad- h(s_t)
\ge -\alpha h(s_t) - \epsilon,\;\;\; \epsilon \ge 0
\end{aligned}
\label{qp}
\end{equation}
All the executed transitions are stored in a replay buffer $B$ to update the RL policy, whereas the tuples $(s_t,\; a_t^{\mathrm{CBF}} + a_t^{bar})$ are stored separately in $\hat B$ to refine the compensator.
The complete online training loop is summarized in Algorithm~\ref{alg:Alg2}.
\begin{figure}[t]
    \centering
    \includegraphics[width=\linewidth]{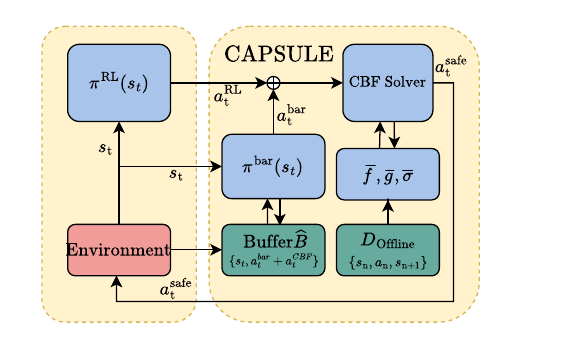}
    \caption{The proposed CAPSULE Algorithmic flow}
    \label{fig:algo_flow}
\end{figure}

\begin{figure*}[t]
  \centering

  \begin{minipage}[b]{0.33\linewidth}
    \centering
    \includegraphics[width=0.8\linewidth]{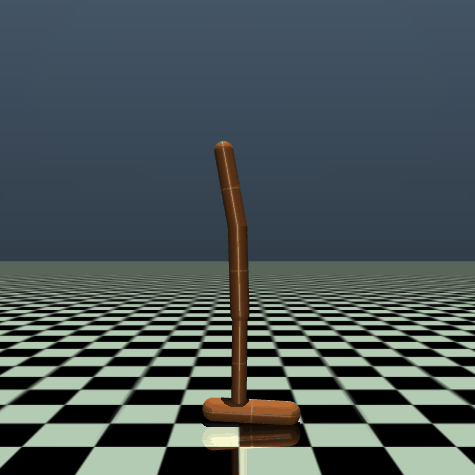}
    \centerline{(a) Hopper}\medskip
  \end{minipage}
  \hfill
  \begin{minipage}[b]{0.33\linewidth}
    \centering
    \includegraphics[width=0.8\linewidth]{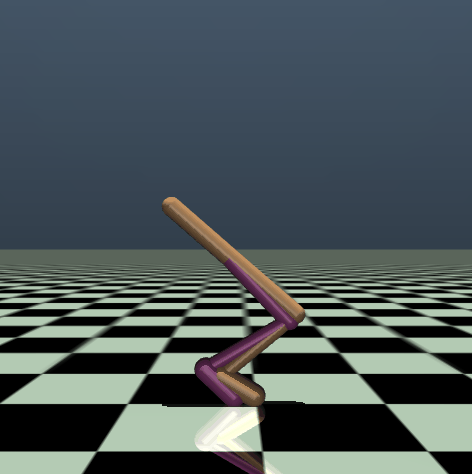}
    \centerline{(b) Walker}\medskip
  \end{minipage}
  \hfill
  \begin{minipage}[b]{0.33\linewidth}
    \centering
    \includegraphics[width=0.8\linewidth]{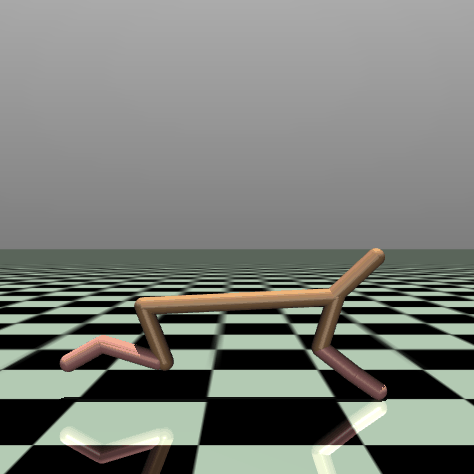}
    \centerline{(c) HalfCheetah}\medskip
  \end{minipage}

  \caption{Visualizations of different MuJoCo control environments used in our experiments: Hopper, Walker, HalfCheetah.}
  \label{fig:benchmark}
\end{figure*}

\section{Experiments}
\begin{figure*}[t]
  \centering

  \begin{minipage}[b]{0.33\linewidth}
    \centering
    \includegraphics[width=\linewidth]{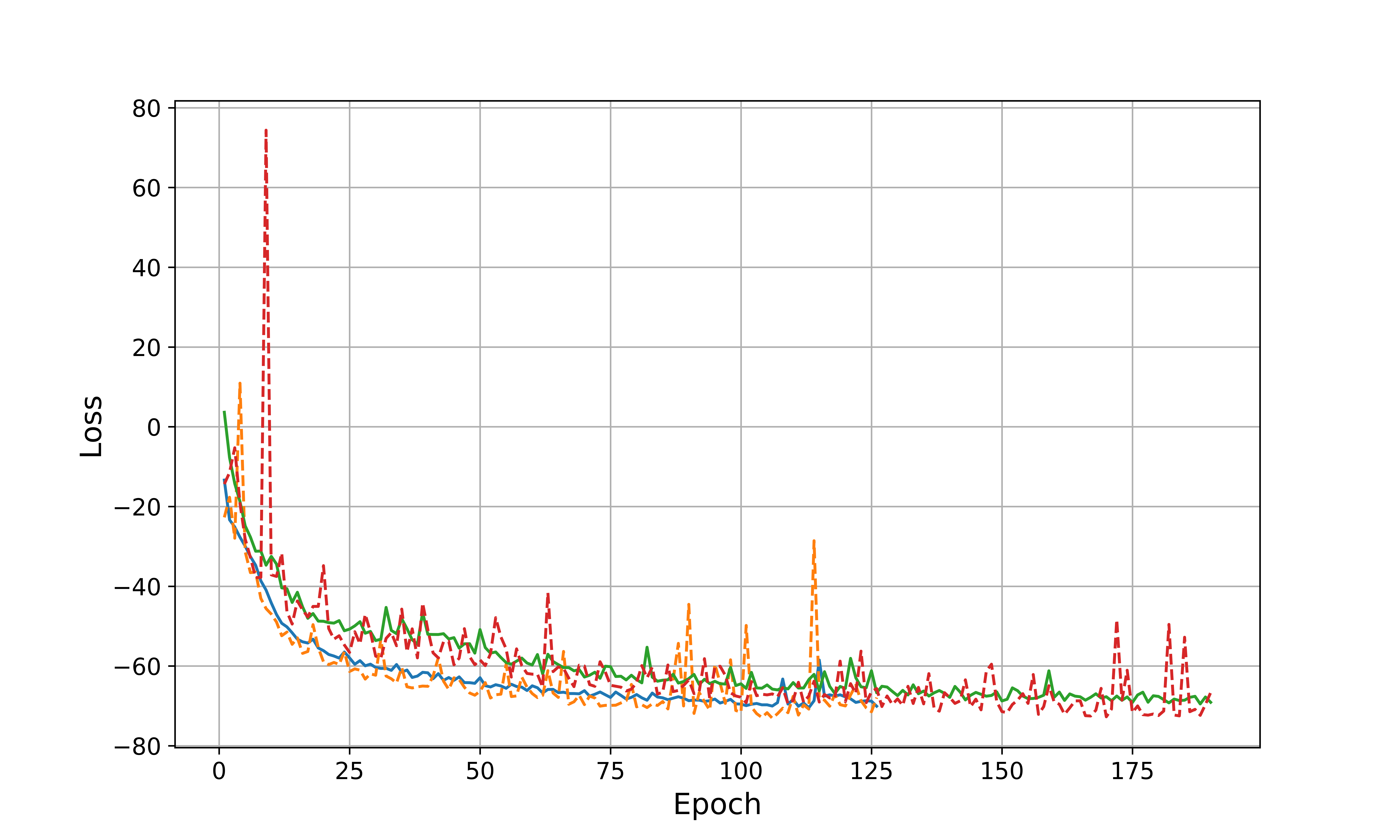}
    \centerline{(a) Hopper}\medskip
  \end{minipage}
  \hfill
  \begin{minipage}[b]{0.33\linewidth}
    \centering
    \includegraphics[width=\linewidth]{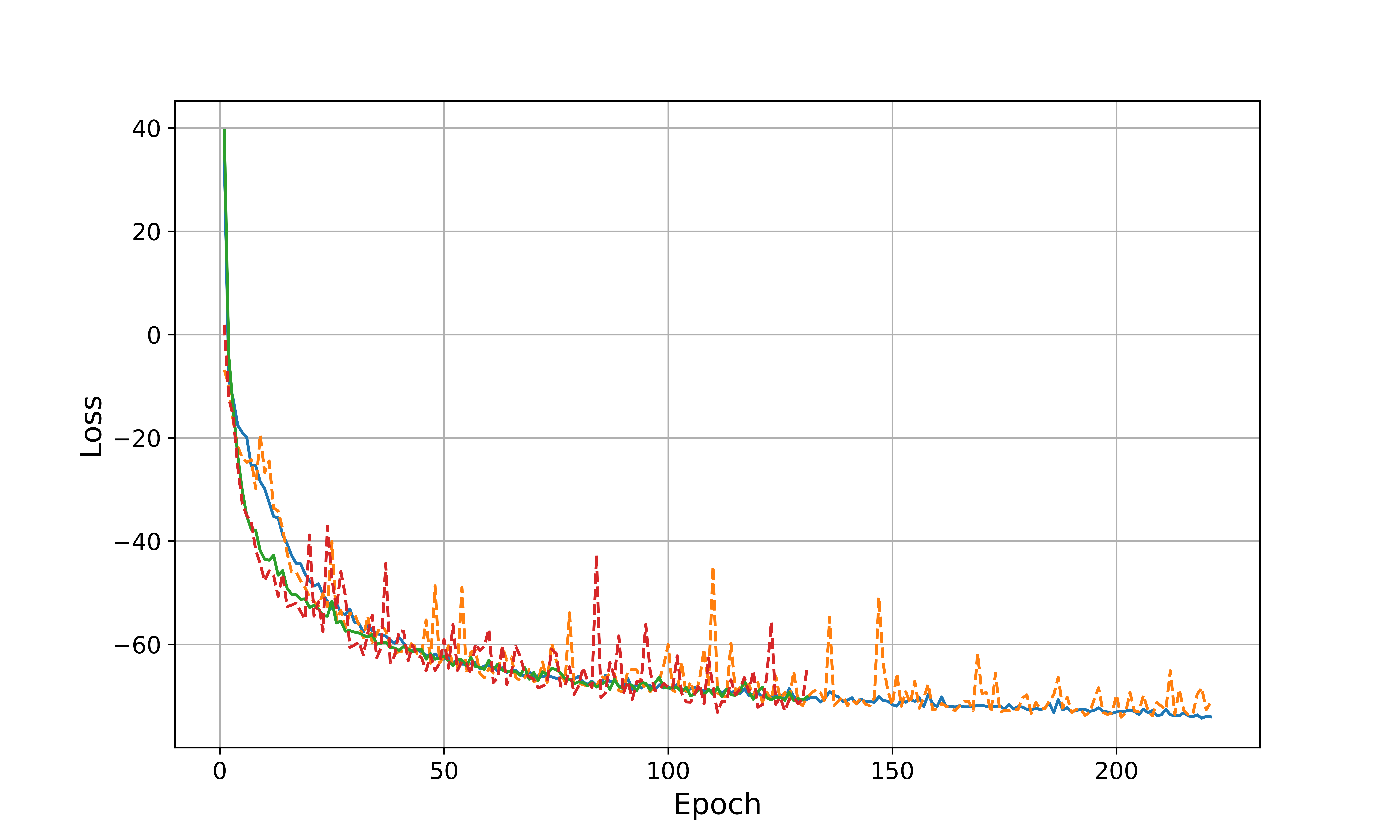}
    \centerline{(b) Walker}\medskip
  \end{minipage}
  \hfill
  \begin{minipage}[b]{0.33\linewidth}
    \centering
    \includegraphics[width=\linewidth]{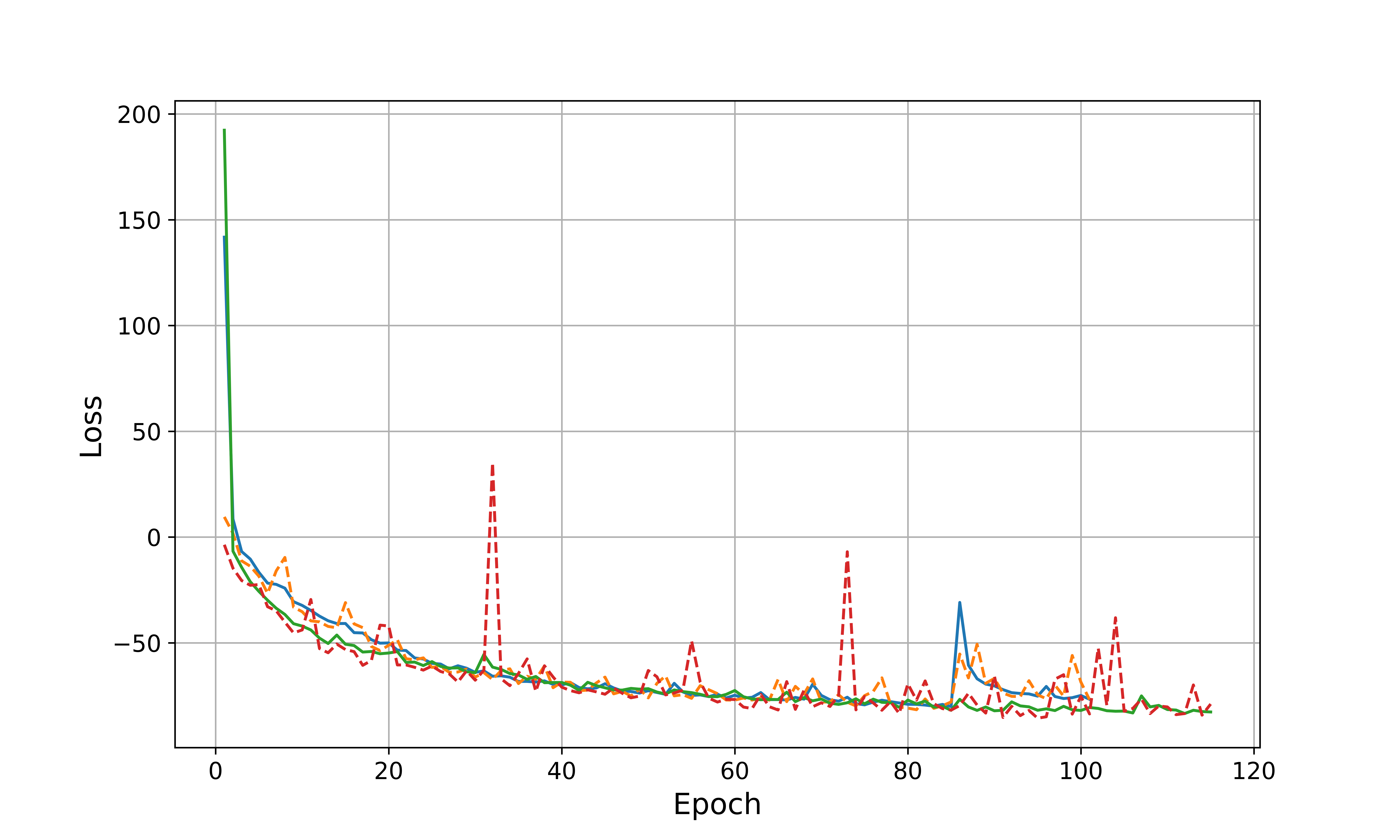}
    \centerline{(c) Halfcheetah}\medskip
  \end{minipage}

\vspace{5pt}
\begin{center}
\begin{tikzpicture}
  \matrix[column sep=15pt] {
    \draw[blue, thick] (0,0) -- (0.6,0); & \node[text=black]{ CAPSULE train loss}; &
    \draw[green, thick] (0,0) -- (0.6,0); & \node[text=black]{    Non linear model train loss}; \\
    \draw[orange, thick, dotted] (0,0) -- (0.6,0); & \node[text=black]{CAPSULE val loss}; &
    \draw[red, thick, dotted] (0,0) -- (0.6,0); & \node[text=black]{Non linear model val loss}; \\
  };
\end{tikzpicture}
\end{center}
  \caption{Pre-training results on MuJoCo continuous control environments.}
  \label{fig:offline_pretrain}
\end{figure*}
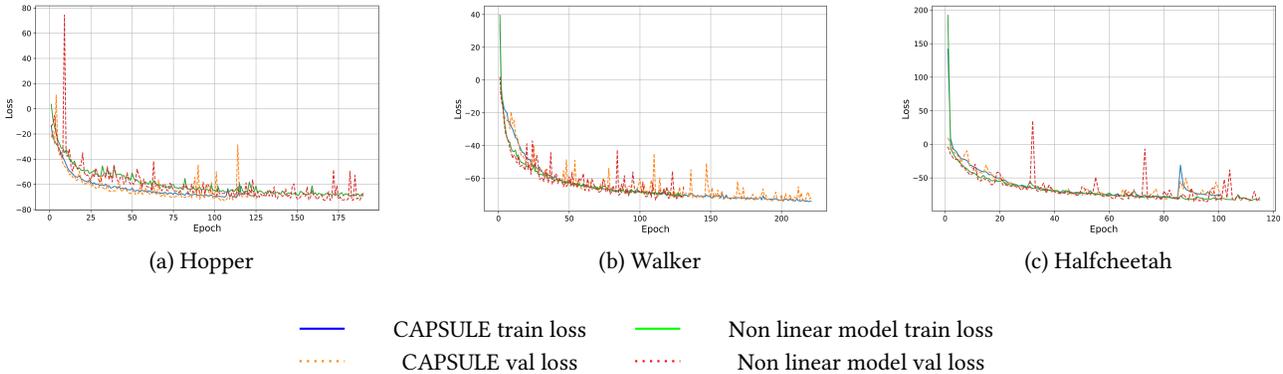

\begin{figure*}[t]
  \centering


    \begin{minipage}[b]{0.95\linewidth}
      \centering

      \begin{minipage}[b]{0.33\linewidth}
        \centering
        \includegraphics[width=\linewidth]{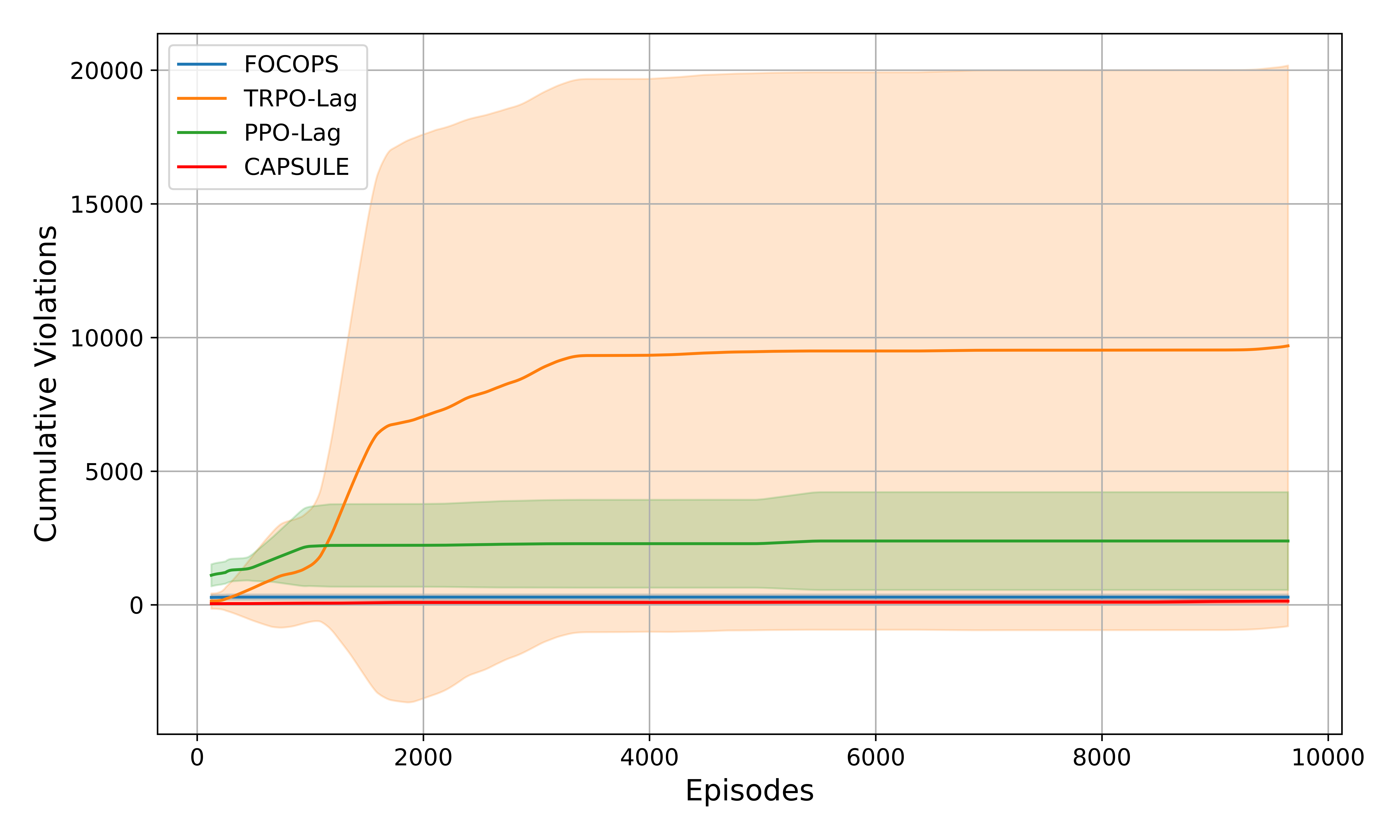}
        \centerline{(a) Hopper}
      \end{minipage}
      \hfill
      \begin{minipage}[b]{0.33\linewidth}
        \centering
        \includegraphics[width=\linewidth]{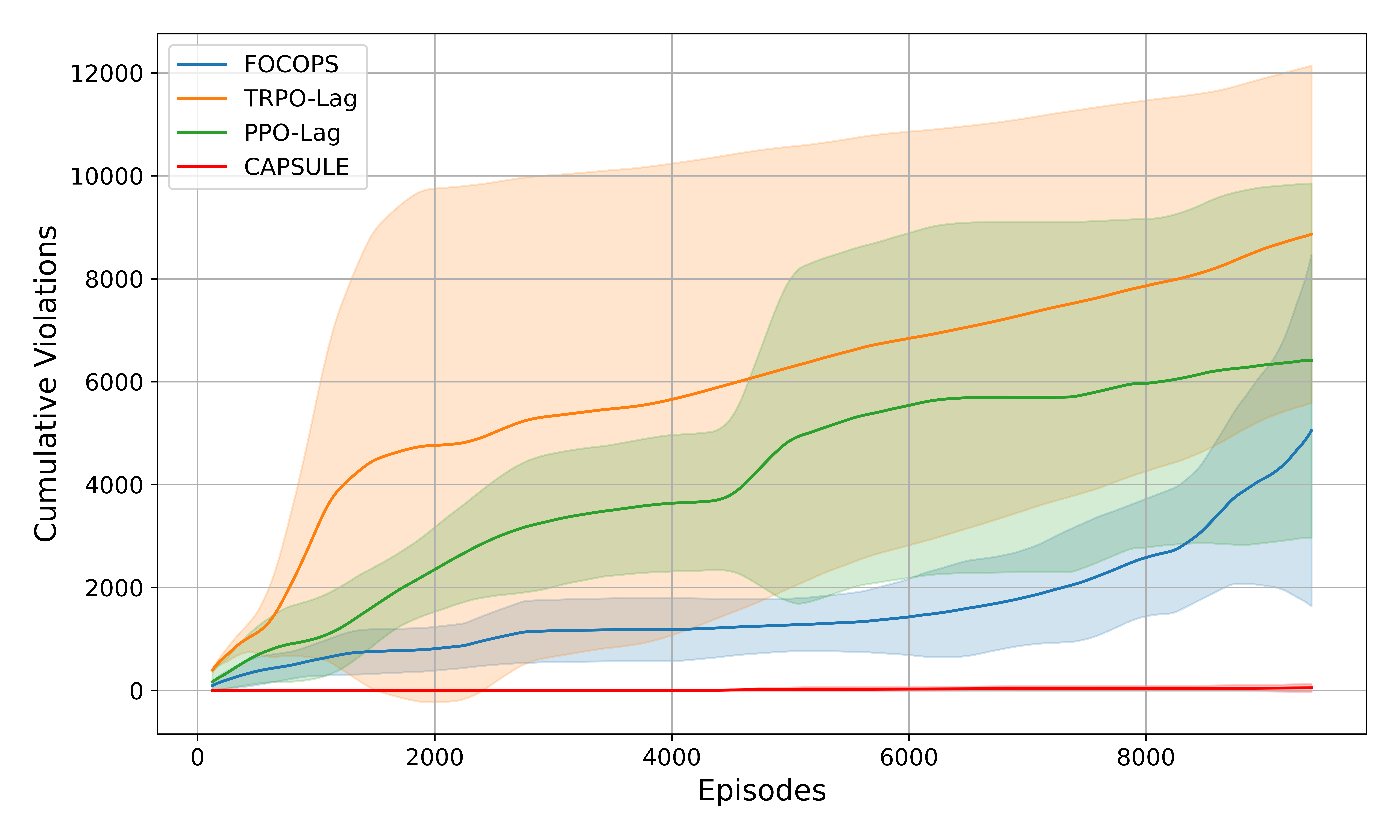}
        \centerline{(b) Walker}
      \end{minipage}
      \hfill
      \begin{minipage}[b]{0.33\linewidth}
        \centering
        \includegraphics[width=\linewidth]{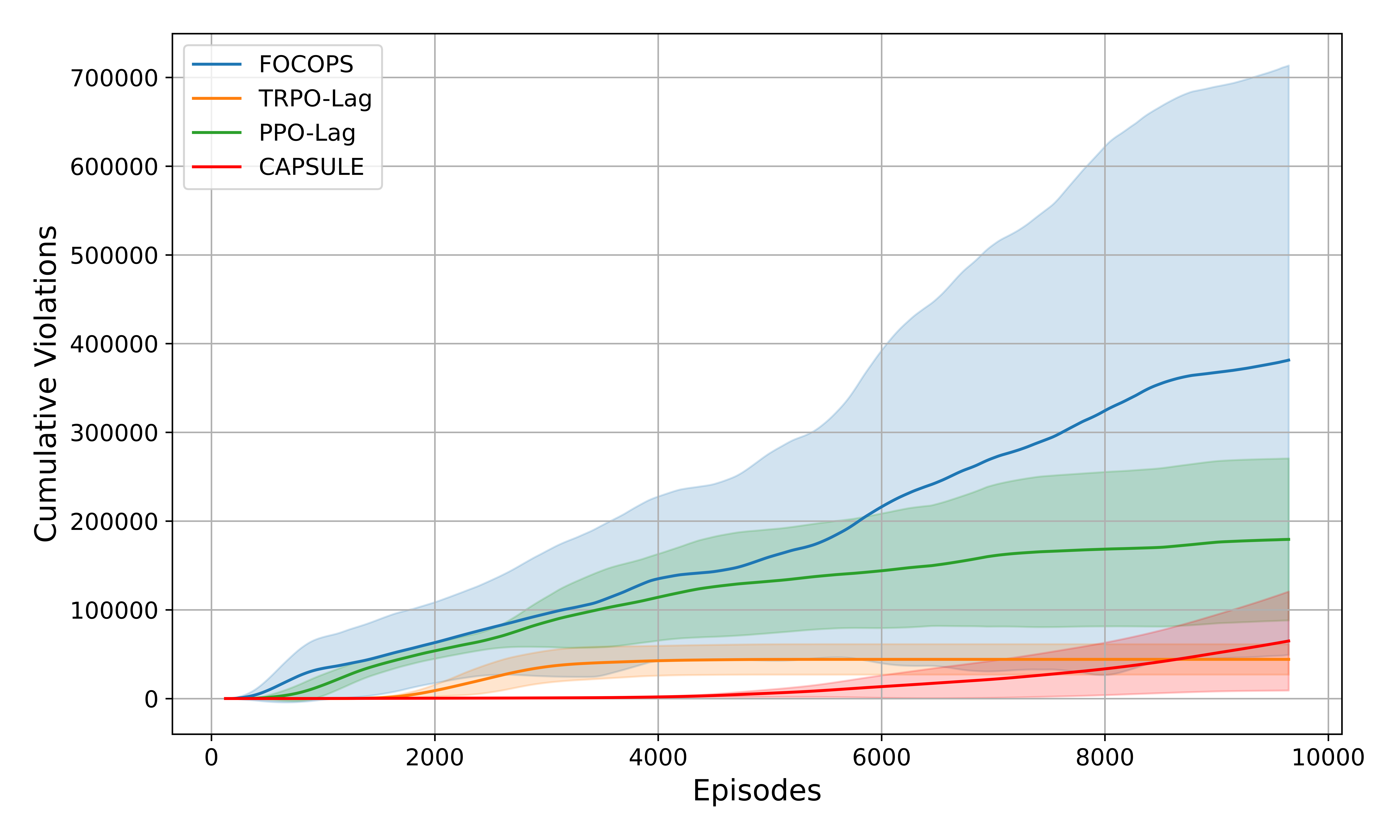}
        \centerline{(c) Halfcheetah}
      \end{minipage}


      \begin{minipage}[b]{0.33\linewidth}
        \centering
        \includegraphics[width=\linewidth]{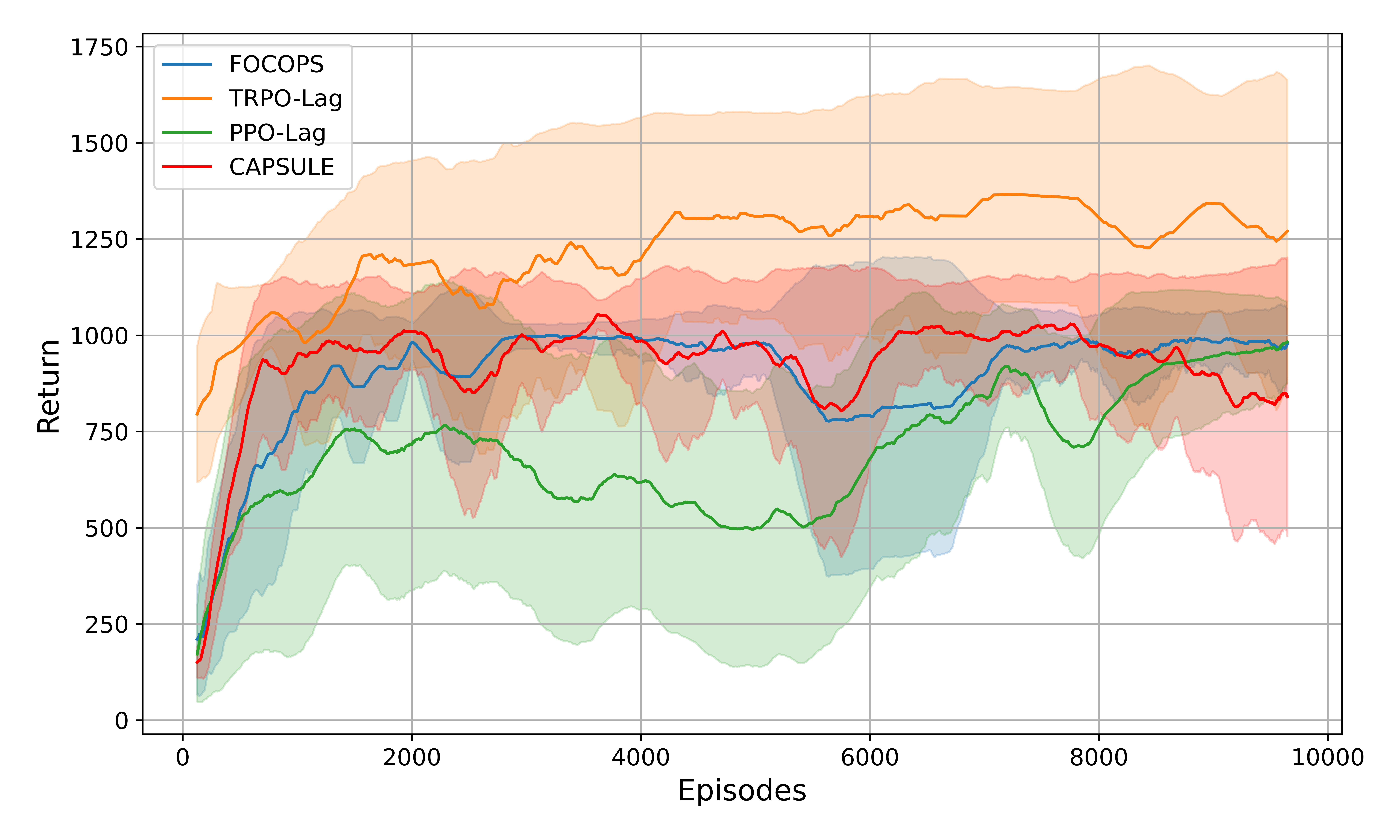}
        \centerline{(d) Hopper}
      \end{minipage}
      \hfill
      \begin{minipage}[b]{0.33\linewidth}
        \centering
        \includegraphics[width=\linewidth]{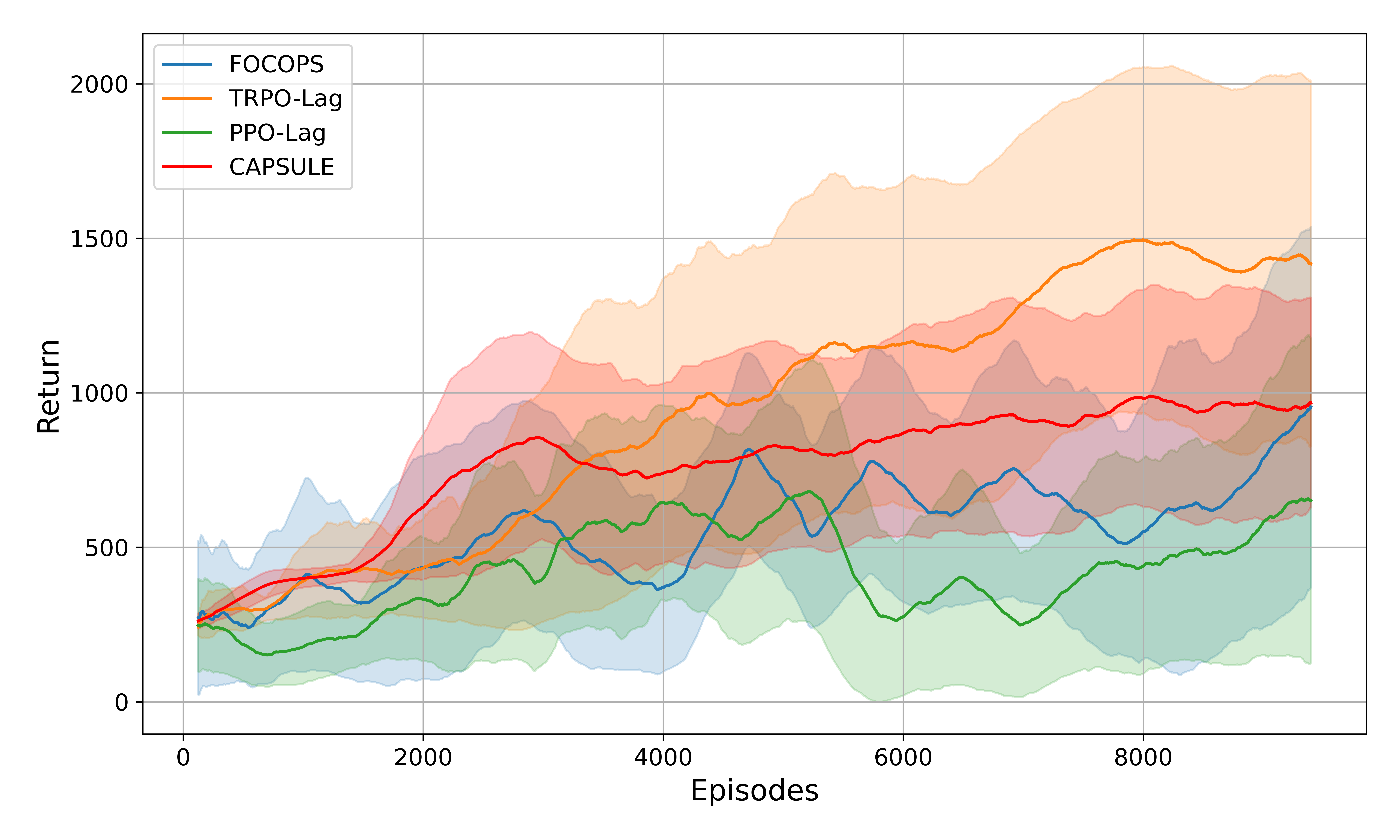}
        \centerline{(e) Walker}
      \end{minipage}
      \hfill
      \begin{minipage}[b]{0.33\linewidth}
        \centering
        \includegraphics[width=\linewidth]{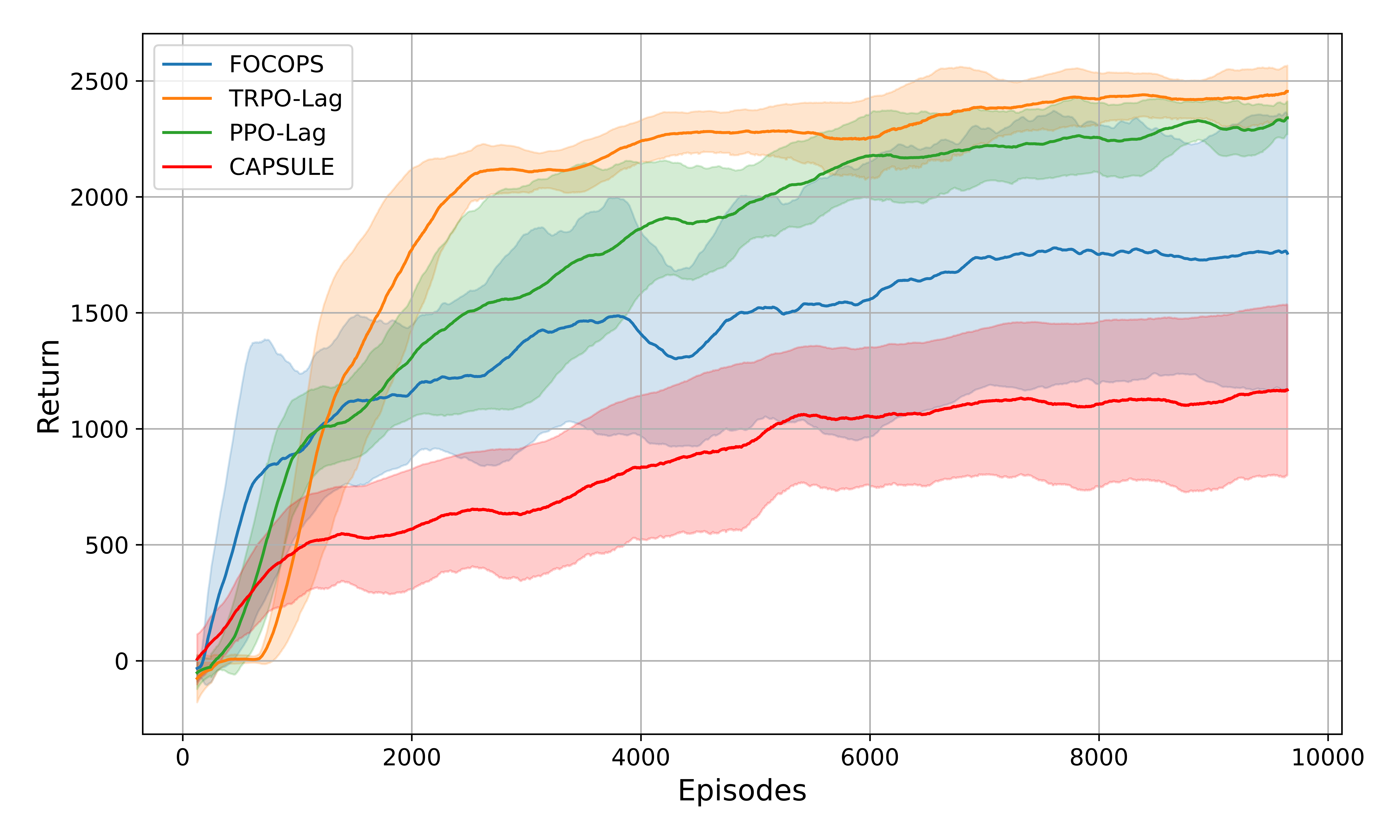}
        \centerline{(f) Halfcheetah}
      \end{minipage}
    \end{minipage}

  \caption{Policy Evaluation on \textit{SafeVelocity} in Mujoco continuos control environments.}
  \label{fig:results}
\end{figure*}
Experiments were conducted in continuous-control environments from MuJoCo which are illustrated in Figure \ref{fig:benchmark}. To obtain a reliable dynamics approximation prior to online learning, we first perform offline pre-training using a dataset of 1M transitions. The corresponding loss curves and learned dynamics accuracy are presented later in the Results section. Comparisons with fully non-linear ensemble model have been given to show we are able to learn good control-affine representations.

After pre-training, the agent is trained online using the proposed CBF-augmented reinforcement learning framework. For empirical evaluation, we compare our method against state-of-the-art Safe Policy Optimization algorithms. We intentionally focus on these CMDP-based baselines because other Safe RL approaches with hard safety constraints typically rely on requirements like known system dynamics, a pre-computed backup (safe) policy, or human interventions during training, and have shown limited success in complex higher-dimensional environments.

We implement these baselines with their safety thresholds modified to $\le0$, \cite{wang}, leading to stricter safety conditions thus leading to a fair comparison between the algorithms.
\subsection{Experimental Setup}

We evaluate our approach under the \textit{SafeVelocity} benchmark introduced in \textit{Safety Gymnasium} \cite{ji2023safety}. In this setting, agents operate in MuJoCo environments and are required to balance task performance with consideration to velocity limits. As described in \cite{ji2023safety}, these tasks encourage agents to move faster while staying alive (not falling, being in range etc.) to obtain higher rewards while simultaneously enforcing safety by assigning a cost whenever a velocity threshold is exceeded.

\subsection{Baselines}
To assess the effectiveness of our method, we benchmark against established safe RL baselines. For fair comparison, all methods use identical reward functions, safety specifications, and environment configurations.
\begin{itemize}
       \item \textbf{TRPO-Lag} \cite{ray2019}: A constrained variant of TRPO that incorporates a Lagrangian dual variable to enforce expected cost constraints while performing trust-region policy updates.

    \item \textbf{FOCOPS} \cite{focops}: A first-order constrained optimization method that stabilizes safe policy learning via a KL-regularized objective and a cost-sensitive update rule.

    \item \textbf{PPO-Lag} \cite{ray2019}: A Lag adaptation of PPO that uses the same dual-variable mechanism as TRPO-Lagrangian but applies it within PPO’s clipped policy gradient framework, making it more computationally efficient.
\end{itemize}
\subsection{Results}
As shown in Figure~\ref{fig:offline_pretrain}, the offline pretraining results across all benchmark environments indicate that our control-affine ensemble model approximates the system dynamics with performance comparable to that of a general nonlinear ensemble model. All the loss curves during training are computed using the loss function defined in \eqref{eq:dyn_loss}.

We trained CAPSULE and all baseline algorithms for $10^6$ environment steps, and all reported metrics are averaged over five random seeds, with variance included to ensure statistical reliability. As shown in Figure~\ref{fig:results}, across all environments our method substantially reduces the number of safety violations.

In the Hopper environment (Figure~\ref{fig:results}(a)), CAPSULE achieves the fewest violations among all baselines, whereas TRPO-Lag exhibits the highest number of violations. As seen in Figure~\ref{fig:results}(d), TRPO-Lag also accumulates higher rewards precisely because it does not adhere to safety constraints, leading to safety violations.

A similar trend appears in the Walker environment (Figure~\ref{fig:results}(b) and~\ref{fig:results}(e)), where unsafe behavior often results in early termination. CAPSULE maintains competitive task performance while drastically lowering violations, capturing the inherent reward–safety trade-off in these environments. TRPO-Lag again demonstrates higher returns at the cost of a significantly increased violation count.

In the HalfCheetah environment, the episode does not terminate when the agent topples or exceeds velocity limits. As a consequence, as shown in Figure~\ref{fig:results}(c) and~\ref{fig:results}(f), agents can accumulate high rewards while simultaneously incurring a large number of violations. This property explains the substantial increase in violations across algorithms. Even in this setting, CAPSULE incurs fewer violations and maintains performance comparable to the baselines.

All presented results are based on evaluation trajectories, with $10$ evaluation episodes generated after each training update.

\section{Related Work}
\subsection{Safe Policy Optimization}
The authors in CPO\cite{cpo} proposed the first policy-gradient methods to address CMDPs in deep reinforcement learning setting. CPO enforces safety constraints while guaranteeing monotonic improvement in expected reward. However, CPO incurs substantial computational overhead due to the need to estimate the Fisher information matrix and optimize a second-order approximation of the objective. Additionally, approximation bias and sampling noise can adversely affect performance. PCPO \cite{pcpo} adopts a two-step strategy that first improves the policy using TRPO~\cite{trpo} and subsequently projects the updated policy onto a constraint-satisfying set to ensure safety. This approach demonstrates improved empirical performance over CPO in certain settings. Nevertheless, PCPO relies on second-order optimization in both stages, resulting in a higher computational cost. On the other hand, FOCOPS~\cite{focops} apply a primal-dual approach to CMDPs by performing policy optimization in a non-parametric space, followed by projection back into the parameterized policy class. While this method is comparatively simple to implement and often exhibits improved sample efficiency, it remains susceptible to instability.
All the above method formulations are based on safety in expectation, and may encounter safety violations. To address this issue, we have proposed a control-theoretic approach for safe RL.
\subsection{Control-theoretic based approaches}
Control Barrier Functions (CBFs) have also been applied to safe reinforcement learning in recent years. Several works aim to reduce or eliminate reliance on known system dynamics by learning safety constraints directly from data. In~\cite{yang2023model}, learns both the policy and the associated CBFs in a data-driven manner, enabling safe policy optimization without access to an explicit environment model. These approaches \cite{marvi2020safe,zhang2024constrained} combine reinforcement learning with CBF-based constraints to ensure safety during learning. \cite{cheng} use CBFs to explicitly restrict the policy search space, thereby improving both safety and sample efficiency, often supported by Gaussian process–based uncertainty estimates. Other works leverage learned or partially known dynamics together with CBFs to provide formal safety guarantees, accompanied by convergence and stability analyses. Empirical results in these studies demonstrate improved performance and stronger safety guarantees compared to baseline safe RL methods. However, all the above methods have been evaluated on simplistic settings and/or assumed to know the system dynamics a priori. We utilize grounded concepts from CBF along with uncertainty-based stochastic models to overcome these issues and evaluate in challenging environments.
\section{Conclusion}
In this work, we introduced CAPSULE, a safe RL framework that employs control-theoretic  formulations to provide safety guarantees and leverages uncertainty-aware dynamics modeling to provide reliable predictions. By learning a control-affine ensemble model offline, we obtain a stable approximation of the control-affine dynamics of the system without requiring access to the true dynamics. This allows the framework to satisfy forward invariance during online interaction. More broadly, CAPSULE illustrates how control-theoretic structures and data-driven uncertainty estimation can be combined to support safe exploration in complex environments.



\bibliographystyle{ACM-Reference-Format} 
\bibliography{sample}


\end{document}